\documentclass[runningheads]{llncs}
\usepackage[T1]{fontenc}
\usepackage{graphicx}
\usepackage{amsmath}
\usepackage{color,soulutf8}
\usepackage{comment}
\usepackage{svg}
\usepackage{verbatim}
\usepackage{float}

\begin{document}
\title{Forecasting Energy Availability in Local Energy Communities via LSTM Federated Learning}
\titlerunning{Forecasting Energy Availability in LEC via LSTM Fed. Learning}
%
\author{Fabio Turazza\inst{1} \and Marcello Pietri\inst{1} \and Natalia Selini Hadjidimitriou\inst{1} \and Marco Mamei\inst{1,2}}
\authorrunning{Turazza et al.}
%
\institute{DISMI, University of Modena and Reggio Emilia, Italy \and
En\&Tech, University of Modena and Reggio Emilia, Italy\\
\email{name.surname@unimore.it}}
\maketitle              
\begin{abstract}
Local Energy Communities are emerging as crucial players in the landscape of sustainable development. A significant challenge for these communities is achieving self-sufficiency through effective management of the balance between energy production and consumption. To meet this challenge, it is essential to develop and implement forecasting models that deliver accurate predictions, which can then be utilized by optimization and planning algorithms. However, the application of forecasting solutions is often hindered by privacy constrains and regulations as the users participating in the Local Energy Community can be (rightfully) reluctant sharing their consumption patterns with others. In this context, the use of Federated Learning (FL) can be a viable solution as it allows to create a forecasting model without the need to share privacy sensitive information among the users. In this study, we demonstrate how FL and long short-term memory (LSTM) networks can be employed to achieve this objective, highlighting the trade-off between data sharing and forecasting accuracy.
\keywords{Local Energy Communities (LECs) \and Federated Learning (FL) \and Internet of Things (IoT) \and Information Sharing \and Forecasting}
\end{abstract}

\section{Introduction}

The increasing vulnerability of urban environments to climate change, alongside their significant contribution to global CO$_{2}$ emissions — exceeding 70\% — ~\cite{UN-Habitat} has positioned cities and local communities at the forefront of the Energy Transition (ET) and efforts toward achieving carbon neutrality~\cite{otamendi2022can,ceglia2020smart}. Given their central role, it is imperative that the ET involving these local communities is executed with a focus on inclusivity and democratic governance, in alignment with the United Nations Sustainable Development Goals outlined in the Agenda 2030~\cite{UN-Agenda2030}. According to the Electricity Market Directive~\cite{EC-COM2016recast}, Local Energy Communities (LECs) are recognized as cooperative structures that manage local energy dynamics, including consumption and distribution. The Renewable Energy Directive further requires that Member States empower citizens to engage in the production, consumption, storage, and trade of renewable energy~\cite{eu-2018-2001}. The 2019 Clean Energy for All Europeans Package~\cite{european2019clean} reinforces the need for a smarter, more efficient energy market, highlighting the critical role of LECs in this evolving landscape and stressing the importance of their recognition as legal entities.

\begin{figure}
\centering\includegraphics[width=0.6\textwidth]{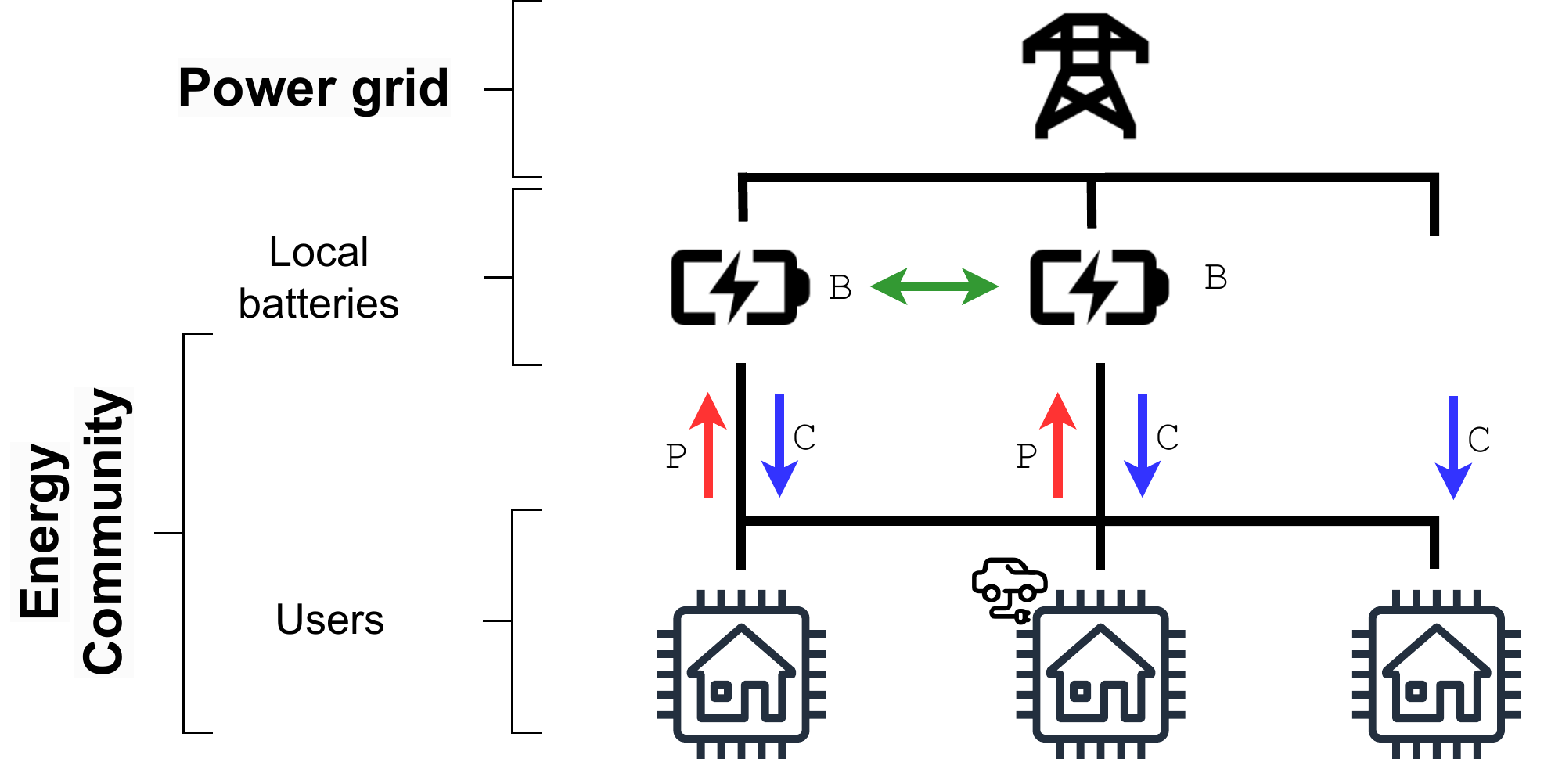}
    \caption{\label{fig:arch} A schematic representation of the Local Energy Community (LEC): multiple users with different energy profiles are connected in a community to share local resources. The goal of the community is to minimize access to the power grid by relying on local batteries (and production).}
\end{figure}

While LECs offer numerous benefits, they also present challenges, particularly in the management of such communities due to rapidly changing regulations and uncertainties regarding how to enhance self-consumption and design the community effectively~\cite{manso2022local}. Figure \ref{fig:arch} presents the LEC architecture: multiple users with diverse energy profiles are connected within a community to share local resources. The community's goal is to minimize reliance on the power grid by utilizing and sharing local batteries and production.

In recent years, there has been a noticeable rise in initiatives related to Energy Communities that incorporate smart grid technologies~\cite{barbour2018community,van2020community}. Smart grids are distinguished by their ability to facilitate bidirectional energy flows, enabling the implementation of demand management strategies and providing the flexibility to dynamically respond to instances of overload. In systems powered by Renewable Energy Sources (RES), consumers can also function as energy suppliers, contributing excess energy back to the grid. These so-called prosumers present numerous opportunities for optimizing energy demand management, particularly in the context of the growing prevalence of electric vehicle (EV) charging. Moreover, prosumers are central to the LECs as defined by the 2019 European Clean Energy Package. The increasing adoption of electromobility has led to a surge in electricity demand, especially in high-density areas such as shopping centers, workplaces, and airports. Within this framework, LECs offer a valuable solution for managing the increasing demand for EV charging.

Accurately forecasting the production and consumption of energy and performing continuous monitoring and management of LECs are crucial tasks, as they would allow to effectively optimize and plan resource usage in the LECs.

It is clear than in order to forecast production and consumption of energy in a LEC it is essential to obtain data about production and consumption of energy from all the members of the community. However, this data acquisition is frequently impeded by privacy concerns and regulatory constraints. Users who participate in these communities may justifiably hesitate to share their energy consumption patterns with others, due to the potential risks to their privacy. This reluctance is rooted in the need to protect sensitive personal data, which can be exposed through the sharing of detailed consumption information. As a result, privacy considerations and existing regulations pose significant challenges to the widespread adoption of these forecasting technologies.

In the context of LECs, Federated Learning (FL) presents a promising solution to the challenge of privacy preservation in energy forecasting. FL enables the development of robust forecasting models without the need for users to share sensitive personal information. Instead of centralizing data, FL allows individual users to collaboratively train a model while keeping their data localized on their own premises. This approach not only addresses privacy concerns but also aligns with increasingly stringent data protection regulations, such as the GDPR (General Data Protection Regulation) in Europe, or the The CCPA (California Consumer Privacy Act) in the United States.

In this study, we explore the application of FL in conjunction with Long Short-Term Memory (LSTM) networks to create accurate energy consumption forecasts. We demonstrate how these techniques can be effectively integrated to maintain user privacy while still providing valuable insights for energy management. Additionally, our research highlights the trade-offs between minimizing data sharing and maintaining high forecasting accuracy, a balance that is crucial for the practical deployment of such systems. By leveraging FL, it is possible to respect user privacy while still harnessing the power of advanced machine learning techniques to enhance the efficiency and sustainability of LECs.

In particular, we present an empirical case study based on an innovative dataset \cite{yuan2023synthetic} to test the aforementioned ideas.

This paper is organised as follows: Section \ref{background} delineates the state of the art in LECs' management and in FL forecasting approaches. Section \ref{lstmfl} describes the presented approach based on LSTM and FL. Section \ref{exp} presents the experimental setup and results. Lastly, Section \ref{conclusions} discusses the results and presents future work.

\section{Literature Review}
\label{background}

In recent years, the development of a digital infrastructure to manage the relationships between nodes within LECs has gained increasing attention in the literature on sustainable solutions for cyber-physical systems. The key substrate of such infrastructure makes forecasting and optimization in the LEC feasible. Due to the complexity of the smart grid and its interactions, several authors have proposed hierarchical models for energy management optimization. In \cite{manshadi2015hierarchical}, the authors presented a model that optimizes the payoff between market participants and distribution network operators. \cite{wu2022hierarchical} developed a system for managing layers within EV charging networks at the microgrid level, while \cite{jiang2022novel} implemented a hierarchical Digital Twin (DT) for smart grid modeling. Similarly, \cite{nagpal2022local} introduced a framework for managing energy in LECs with Community Energy Storage (CES), where individual buildings optimize energy costs and the CES coordinates to maximize self-consumption. The concept of Peer-to-Peer (P2P) energy sharing, which involves prosumers in the energy market for the commercialization of excess energy, is vital in promoting the use of renewable energy sources \cite{soto2021peer}. However, the real-time management of energy exchanges among users demands advanced digital platforms and smart metering systems \cite{piselli2021assessing}. Furthermore, reducing central energy consumption, especially with the introduction of EVs and energy storage systems, requires effective management strategies. 

In this context, our forecasting model, based on FL, can leverage such an infrastructure like the ones cited above to manage the exchange of excess energy within an LEC comprising prosumers with energy storage systems and EVs, as well as consumers. This infrastructure not only will facilitate energy exchange but also necessitates and supports the implementation of energy forecasting mechanisms for the optimal operation of LECs.

Traditional ML Centralized models merge different datasets in a single one. This approach risks sensitive data leakage \cite{Zhang2022}. Therefore, FL, proposed by \cite{McMahan2016}, trains locally models without sharing raw data. 
FL applied to the energy sector is an innovative approach that allows for the development of collaborative models without the need to disclose sensitive data. This feature is particularly important in the energy sector, where the data collected is almost always Cross-Silo and distributed among various consumer entities that are unaware of the predictive model's training process. However, this approach presents several challenges, including the quality control of local data and the management of its heterogeneity in a Centralized context. Although this problem is usually addressed with scrupulous preprocessing of the available data or during the data merging phase through complex aggregation algorithms such as FedProx and FedMa~\cite{ENERGYeFL_9949863}. Another significant challenge concerns the synchronization of models with data that can vary in their update frequency, ranging from sporadic real-time usage, and the optimization of communication between nodes, with a high volume of transmitted data and the complexity of predictive models in forecasting which potentially causes significant network overhead~\cite{Edge-Based_Communication_Optimization}. In an energy context, it is also important to consider the variability and instability of the data, especially in domains such as solar and wind energy, which are highly unpredictable; therefore, the final model must have a robust and accurate structure capable of adapting to the instability and disturbances typical of the energy domain.

In the LECs' scenario, data privacy presents a significant challenge when training data-driven models, as customer data such as consumption profiles is often required. Customers are typically hesitant to share data with Centralized servers due to privacy concerns, commercial competition, and technical barriers. Additionally, traditional Centralized model training involves substantial computing resources and exposes data transmission to cyberattacks, adding to network overhead \cite{ENERGYeFL_9949863,ENERGYeFL_9548947}.

Despite acknowledging the benefits of FL in preserving privacy and reducing data transmission, recent works like \cite{ENERGYeLEC_noFL_PUTZ2023100983} focus on optimizing energy management in LECs by evaluating the quality and value of various forecasting methods, integrating them into a Model Predictive Control (MPC) framework to improve performance indicators such as load cover factor and energy costs. However, unlike our approach, these studies do not employ FL, which can often lead to variations in results — sometimes worsening — due to the lack of decentralized data processing and the potential privacy risks associated with Centralized data collection.

FL provides a viable solution by allowing distributed model training across multiple clients using their local data. A central server coordinates the process, aggregating the locally trained models into a global model, thus preserving data privacy by keeping raw data on local devices and minimizing network overhead, as only model updates are shared. FL has gained popularity in fields such as Internet of things (IoT) and Edge Computing, especially for privacy protection. Although its application in the energy sector is still emerging, interest is growing rapidly as FL's advantages extend beyond just privacy concerns \cite{ENERGYeFL_9949863}.

In \cite{ENERGYeFL_9469923}, the authors propose a FL architecture combined with Edge Computing for short-term residential energy consumption forecasting, using LSTM models trained locally on user devices. The approach enhances privacy by keeping data local, and incorporates weather and calendar features, with user clustering based on consumption or socioeconomic similarities to improve forecasting performance.
In \cite{ENERGYeFL_9548947}, the authors propose a privacy-preserving method for disaggregating com\-munity-level behind-the-meter (BTM) solar generation using a FL-based Bayesian Neural Network (FL-BNN). This approach allows utilities to train models locally, preserving data privacy while accurately estimating solar Photovoltaic (PV) generation by capturing uncertainties through a layerwise parameter aggregation strategy.

Our work expands both \cite{ENERGYeFL_9469923} and \cite{ENERGYeFL_9548947} by applying FL to a broader energy management framework within LECs. We incorporate multiple energy sources presenting an empirical case study based on an innovative dataset \cite{yuan2023synthetic}, using advanced forecasting methods like LSTM networks. Additionally, we develop specialized aggregation algorithms that enhance prediction accuracy and adapt to the diverse needs of LECs, all while preserving data privacy.

\section{LSTM Federated Learning}
\label{lstmfl}


As described in \cite{Wen2023}, FL takes into account $N$ participants, i.e. members of the LEC that want to leverage trained model. Users merge their knowledges $\{p_1, p_2, \dots, p_N\}$ to train the model. The FL model is trained by minimizing the loss function in Eq. \ref{fl_loss}. Where $n_k$ is the amount of data on the member $k$ and $F_k(w)$ is the local objective function. 
\begin{align}
    \min f(w) = \sum_{k=1}^N \frac{n_k}{n} F_k(w) \label{fl_loss}
\end{align}
The global model parameters are set up in a central server. Each client download the global parameters, trains its own model, and finally updates the global model round after round \cite{Zheng2023}. FL can be divided in three primary categorizations: horizontal FL, vertical FL, and Federated transfer learning. Horizontal type deals with overlapping features among the datasets of its members. In contrast, vertical one deals with overlapping users with less overlapping features. The last type rarely has overlapping datasets but it can overcome data scarcity by transferring learning \cite{Zhang2021}. 

In this work we focus on horizontal FL as we have multiple members of the LEC having datasets exhibiting overlapping features  (i.e., energy consumption and production patterns for each member).



To make predictions in forecasting, we employed a Long Short-Term Memory Network (LSTM) given its ability to model the multiple time-scales patterns associated to energy consumption and production \cite{greff2016lstm}.

Combining Long Short-Term Memory (LSTM) networks with FL offers a powerful approach for handling sequential data while preserving data privacy with some key advantages such as the Reduced Risk of Overfitting with the aggregation of updates from multiple clients with potentially diverse datasets and the reduced communication costs by processing data locally on devices, sharing only model updates, making the FL suitable for Real-Time predictions, even though this aspect will not be considered in this paper.

The method we propose to address this scenario is a Model-Centric Cross-Silo Horizontal architecture including several entities that represent individual prosumer or consumer users. These entities perform Client-Side Forecasting of the net energy production on an hourly basis.

\begin{align}
Net\_Energy _{x} = \sum_{x} E_{exp,x} - \sum_{x} E_{imp,x}
\end{align}

\begin{itemize}
    \item \( E_{exp,x} \) = Total Net Energy from User \(x\)
    \item \( E_{exp,x} \) = Exported\_energy from User \(x\)
    \item \( E_{imp,x} \) = Imported\_energy from User \(x\)
    \item \( x \) = User Index (Prosumer or Consumer)
\end{itemize}

\begin{figure}
    \centering
    \includegraphics[width=0.9\textwidth]{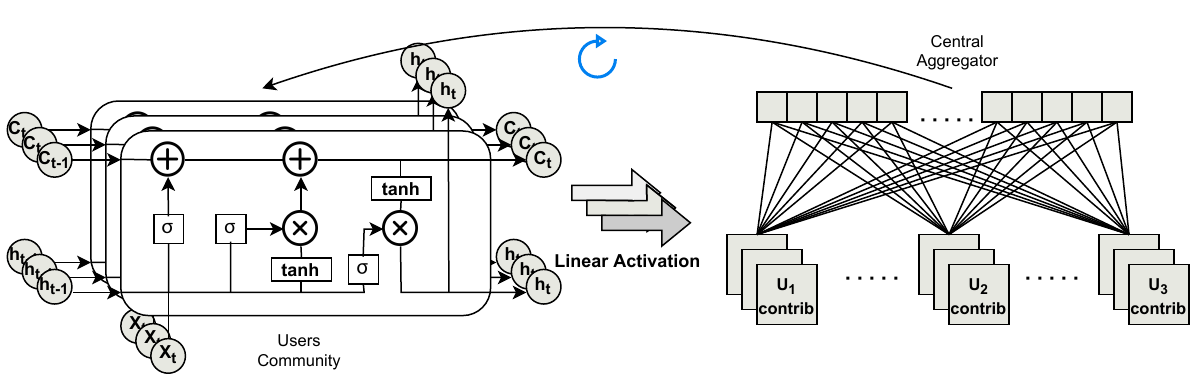}
    \caption{\label{fig:LSTMFEDarch} This is our proposed FL architecture: on the left, there are several LSTM layers that represent different client instances, also called 'runners'. The clients train on their local data using a model shared by the server and send the parameters to it. The server then acts as a central aggregator and combines the parameters to obtain a better global model.}
\end{figure}

Subsequently, the parameters of the models containing the temporal relationships with a 24-hour dependency are aggregated Server-Side through a custom extension of Federated Averaging (FedAvg). This effectively creates a new dataset of parameters which, while maintaining the temporal relationships of the different clients, will be sent back to each client for a pre-defined number of federated epochs.
At the end of Federated Learning, the predictions are centrally associated with the reference time instant through a FedProx aggregation algorithm and they are summed by filtering the frames for each index. In this way, we move away from the concept of individual user energy to talk about the net energy of the LEC community. A synthetic architecture overview of the Federated flow is shown in Figure \ref{fig:LSTMFEDarch}.

\section{Experimental Setup and Evaluation}
\label{exp}

This Section outlines the dataset and the experimental setup used in our study and presents the evaluation of our experiments, including the interpretation of results supported by visual representations.

\subsection{Dataset and Experimental Setup}
\label{dataset_setup}

In this study, we conducted a series of experiments focused on predicting electricity surplus availability within a LEC. Our aim is to assess a range of LEC configurations, including varying percentage of prosumers and consumers in order to study recurring patterns within these communities.

We created a dataset of 200 prosumers and 200 consumers based on the synthetic data provided in \cite{yuan2023synthetic}. This is a dataset created by using real consumption data of Danish residents. The dataset consists of hourly energy production and consumption on weekends/holidays and weekdays in the four seasons. The dataset also contains infraction about the average hourly temperature. On this basis, we created our dataset by randomly joining consecutive days and by later applying a moving average to obtain smoother time series.
We applied the exponential weighted moving average with a window on 120 hours (5 days).
%

Given this dataset, we applied our LSTM-FL to forecast energy surplus availability of multiple prosumers within a LEC. The LSTM is configured as Stateless and receives as input the features from a single user such as historical energy flows in a year, time of the day, and temperature. Furthermore, the model generates a 24-hour energy forecast for each user. The training process learns the surplus behaviour from multiple similar users within the LEC. Once the network is trained, we will be able to predict the energy output generated by each user for each period of the year. Summing all the predictions, filtered by year and time slot, it allows us to derive the energy surplus available to the community (LEC).

\subsection{Evaluation}
\label{evaluation}

\begin{figure}
\centering
A) \includegraphics[width=0.45\textwidth,height=5.5cm]{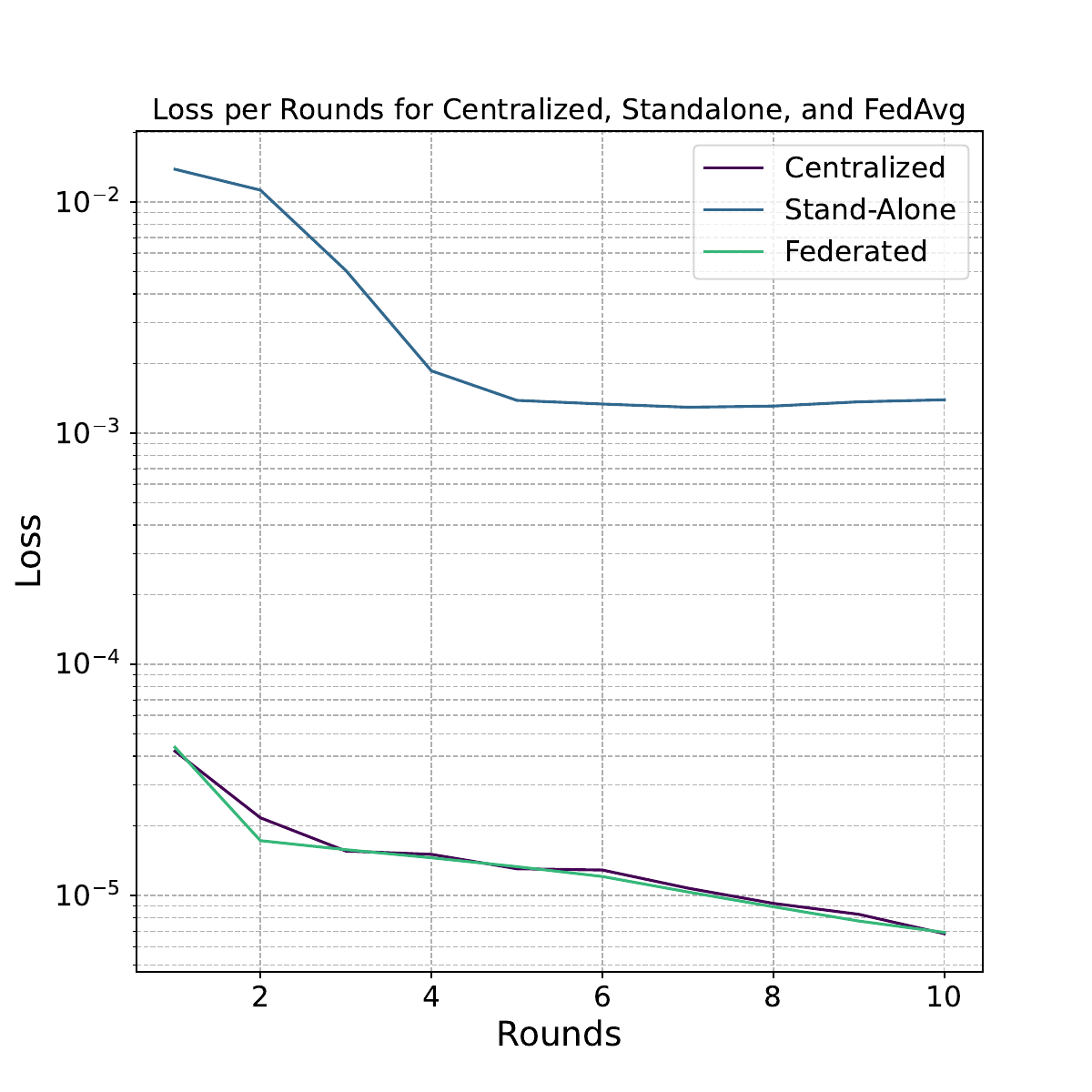}
B) \includegraphics[width=0.45\textwidth,height=5.5cm]{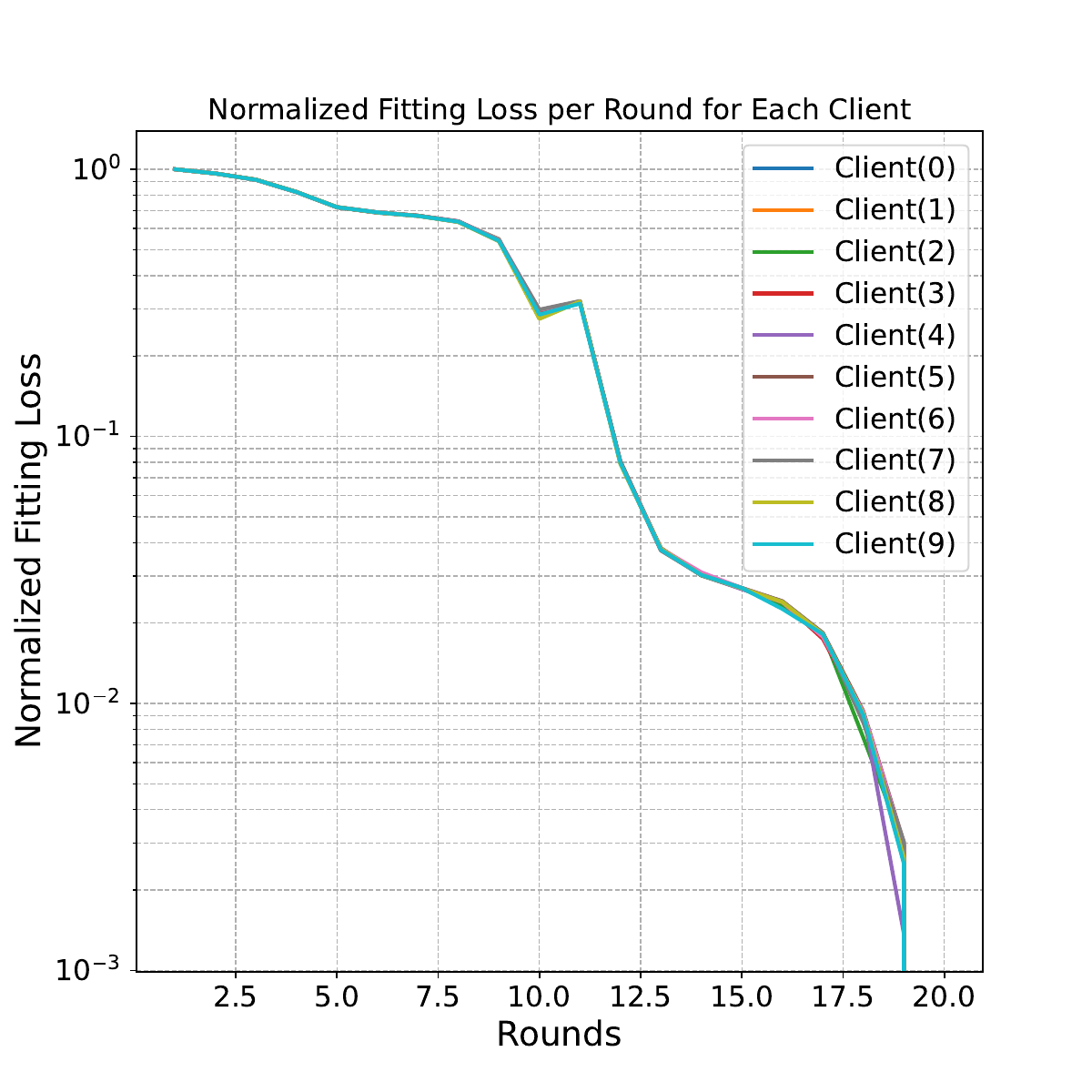}
\caption{\textbf{A)} Mean Squared Error (MSE) of the three approaches (i.e., Stand-Alone, Federated, Centralized learning), considered a small subset of 10 users and a 10 epochs training (for FL they are Federated Rounds). The plot shows the mean MSE of the different stores during testing phase. \textbf{B)} 10-clients LEC MSE fitting losses performance comparison over time, after a flow of 10 Federated Rounds.}
\label{three_scenarios}
\end{figure}

In our experimental evaluation, the first analysis focuses on surplus energy prediction across the three distinct scenarios, including a stand-alone model (thus considering single-user communities), a Centralized model trained on a dataset formed by concatenating all the raw data from a LEC, and finally, a Federated model trained on the same data as the Centralized model but without the explicit sharing of data within the LEC. The following line-charts (Fig. \ref{three_scenarios}-A and Fig. \ref{three_scenarios}-B) show the MSE loss obtained in the different approaches considering a small subset of 10 users. In Figure \ref{three_scenarios}-B, we can observe the actual homogeneity in the training loss of the various clients in the Federated approach. Every LSTM model was trained using a 3-layers LSTM net with 50 neurons as hidden\_size parameter, we used Adam as optimizer algorithm. The models were trained on 8 features:

\begin{itemize}
    \item \texttt{day} (\texttt{int}): The day of the month, represented as an integer (0-335).
    \item \texttt{index} (\texttt{int}): The daily hours, represented as an integer (0-24).
    \item \texttt{temperature\_median} (\texttt{float}): The median temperature recorded, represented as a floating-point number.
    \item \texttt{temperature\_std} (\texttt{float}): The standard deviation of the temperature, represented as a floating-point number.
    \item \texttt{season} (\texttt{str}): The season of the year (e.g., winter, spring), represented as a string.
    \item \texttt{type} (\texttt{Bool}): A boolean value indicating the type of User (Prosumer = 0, Consumer = 1).
    \item \texttt{type2} (\texttt{int}): An additional type represented as an integer indicating a subset type of Users and what kind of energy they produce and consume.
    \item \texttt{sum\_total\_ewm\_balance} (\texttt{float - target}): The total sum of the balance per hour (energy surplus), represented as a floating-point number.
\end{itemize}

As illustrated in Figure \ref{three_scenarios}-A, It is evident that performance improves drastically when moving from a Stand-Alone architecture to a Centralized or Federated one. Between these latter two architectures, we naturally observe slightly better performance in the Centralized model compared to the Federated one, which, however, preserves the privacy of each individual user's data.

\begin{figure} 
    \centering
    \includegraphics[width=0.9\textwidth]{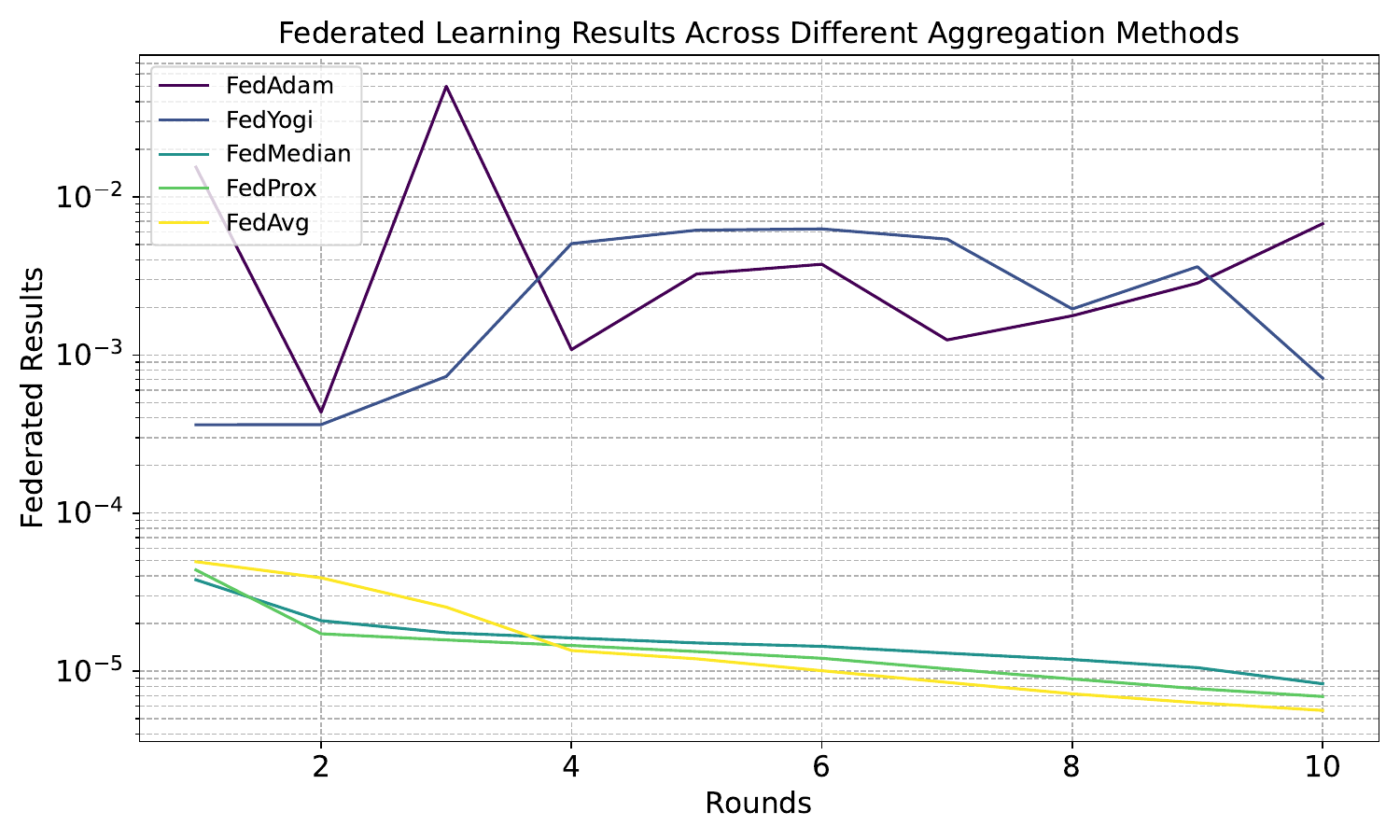}
    \caption{\label{five_methods} Federated aggregation methods MSE loss comparison including five of the most used aggregation algorithms such as FedAvg, FedMedian, FedProx, FedAdam and FedYogi on a 10-users small subset, these methods were tested in 10 Federated rounds.}
\end{figure}

For the Federated experiment, the performance was compared using five of the most famous different aggregation methods (shown in Fig.\ref{five_methods}) including FedAvg, FedMedian, FedProx, FedAdam and FedYogi.

FedProx \cite{DBLP:journals/corr/abs-1812-06127} was chosen for this experiment due to its effectiveness in handling non-IID (Non-Independent and Identically Distributed) data, which is common in FLscenarios. Unlike FedAvg (that combines the local model updates from multiple clients by simply averaging them), which assumes that all clients' data is identically distributed, FedProx, or Federated Proximal, introduces a proximal term ($\mu = 0,01$) to the traditional framework. This term is designed to tackle the challenges posed by heterogeneous data distributions among users. By penalizing deviations of local updates from the global model, FedProx maintains stability and accelerates convergence ensuring that the local updates remain closer to the global model, which enhances the efficiency of the training process and contributes to faster convergence.

The objective function of the FedProx algorithm in the context of FL is defined as follows:

\begin{equation}
\min_{w} \left\{ \frac{1}{K} \sum_{k=1}^{K} \left( \frac{1}{n_k} \sum_{i=1}^{n_k} f_{k,i}(w) \right) + \frac{\mu}{2} \| w - w^t \|^2 \right\}
\end{equation}

Where:
\begin{itemize}
    \item \(\min_{w}\) indicates that we are minimizing with respect to the global weight vector \(w\).
    \item \(\frac{1}{K} \sum_{k=1}^{K}\) represents the average over all clients.
    \item \(\frac{1}{n_k} \sum_{i=1}^{n_k} f_{k,i}(w)\) represents the average of the losses over the samples from each client \(k\).
    \item \(\frac{\mu}{2} \| w - w^t \|^2\) is the Prox regularization term, which penalizes deviations from the global weights \(w^t\) of the current iteration.
\end{itemize}

The term \(\frac{1}{K} \sum_{k=1}^{K} \left( \frac{1}{n_k} \sum_{i=1}^{n_k} f_{k,i}(w) \right)\) represents the average loss over data from different clients, while the term \(\frac{\mu}{2} \| w - w^t \|^2\) acts as a Prox regularization term that helps stabilize the optimization by penalizing deviations from the previous iteration's global weights.

\begin{figure}[t]
    \centering
    \begin{minipage}[b]{\textwidth}
        \centering
        \includegraphics[width=1.0\textwidth]{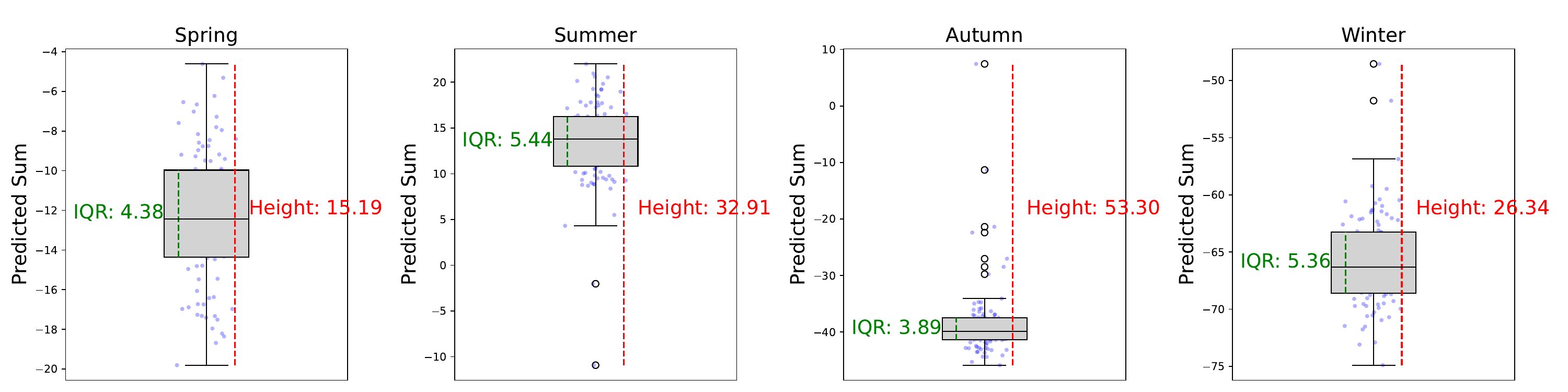}
        \caption{Seasonal Box-Plot of Mixed LEC which shows the interquartile ranges, whiskers, and outliers of a 50/50 LEC.}
        \label{boxplothorizontal5050}
    \end{minipage}
    
    \vspace{0.5cm} 
    
    \begin{minipage}[b]{\textwidth}
        \centering
        \includegraphics[width=1.0\textwidth]{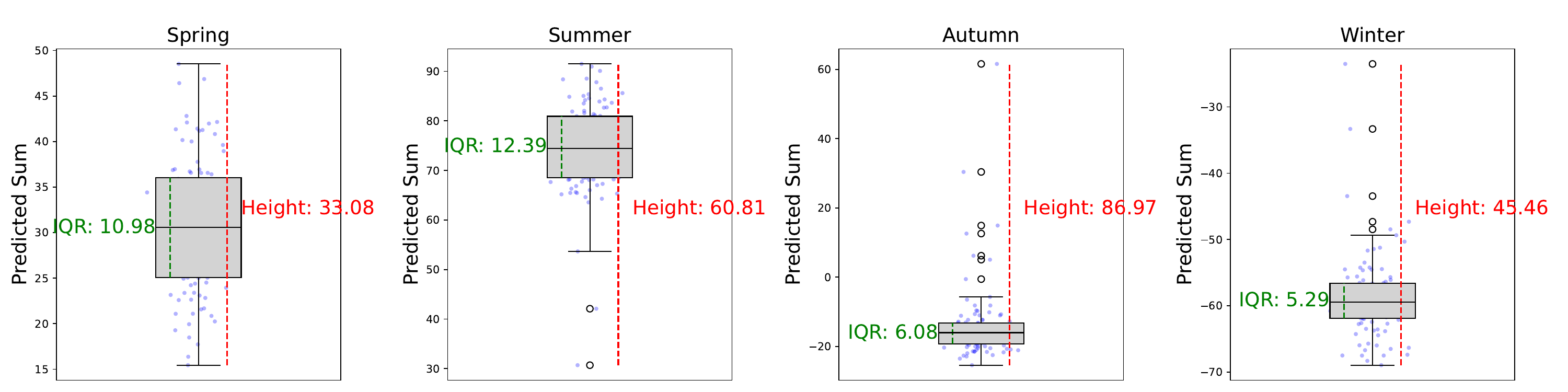}
        \caption{Seasonal Box-Plot of FPL which shows the interquartile ranges, whiskers, and outliers of a FPL.}
        \label{boxplothorizontal1000}
    \end{minipage}
\end{figure}

\begin{figure}
    \centering
    \includegraphics[width=0.95\textwidth]{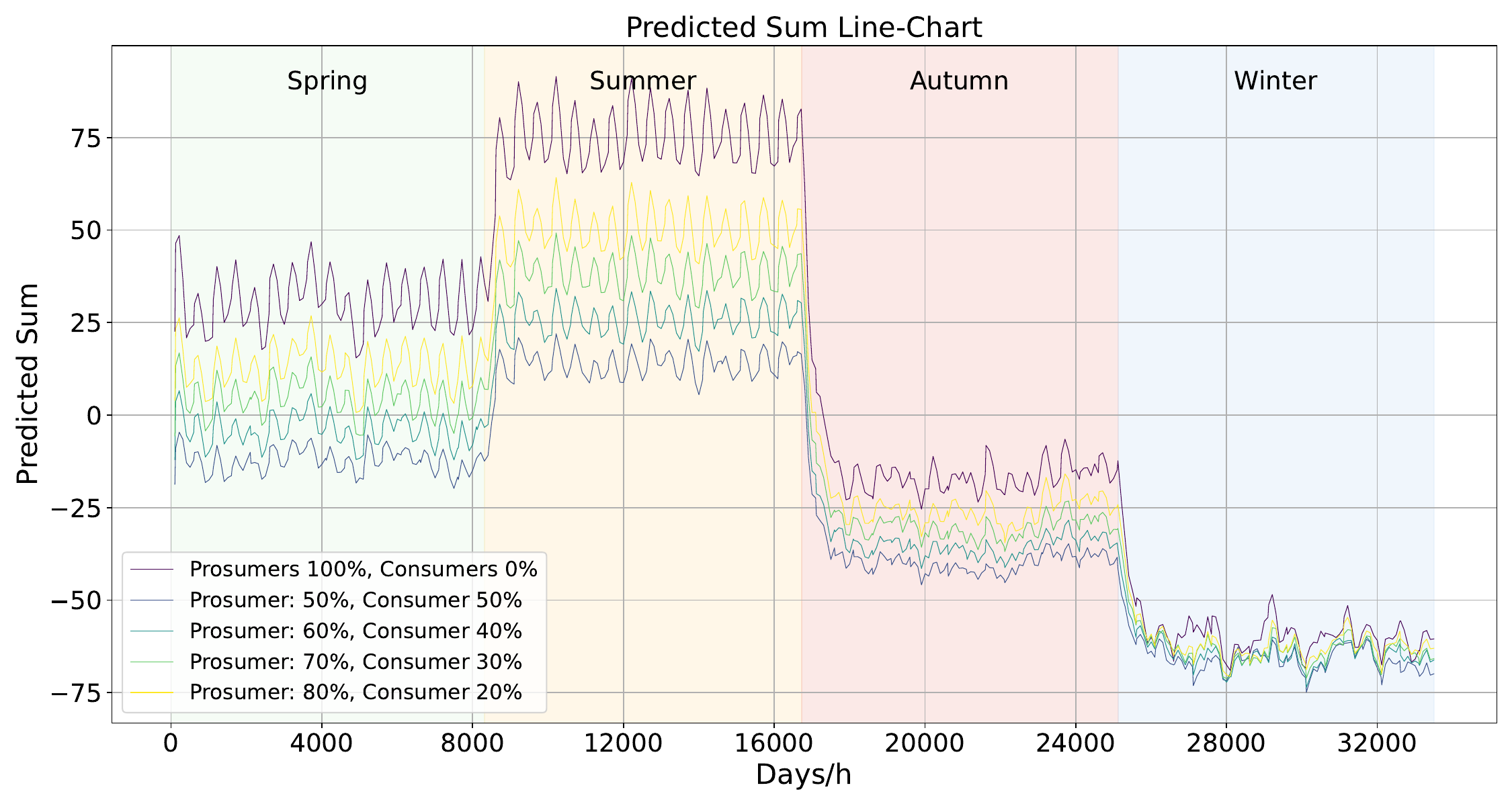}
    \caption{\label{Comparison between mixed LECs} Comparison between different mixed communities, divided by percentages of consumers. The following values are derived from tests conducted in homogeneous communities (with more or less the same percentage of user types) and they are all performed 100 users LEC.}
\end{figure}

\subsubsection{Mixed LEC community}
\label{Mixed LEC community}



In a mixed LEC, prosumers produce and supply excess energy to consumers. Energy overproduction, leading to community autonomy, often occurs in spring and early summer due to high outdoor activity and abundant photovoltaic generation.
The impact of consumers in a community (mixed LEC) is primarily observed during warm seasons, a period in which the variability (measured by the difference between the third quartile ($Q3$) and the first quartile ($Q1$)) among inter-LEC values increases by more than 150\% in the spring months, reaching an increase of only 1.32\% in the winter values. This data makes forecasting the warm seasons more uncertain, encompassing a much larger range of values. This trend is not reflected uniformly in the length of the whiskers in the box-plot, which remains fairly consistent across seasons, excluding any potential outliers.
The interquartile range metrics through box-plots were obtained by comparing a homogeneous mixed LEC (Fig. \ref{boxplothorizontal5050}) (50\% prosumers, 50\% consumers) with a Full Prosumers LEC (FPL) (Fig. \ref{boxplothorizontal1000}).

In the line chart above (as shown in Fig.\ref{Comparison between mixed LECs}), the actual final prediction of the Federated model after a 10-round session in a mixed community is represented. The model was trained on a dataset of 257 users, with hourly tracking, evenly split between prosumers and consumers. The same model was tested on an equivalent dataset of 100 users, divided into 50\% prosumers and 50\% consumers, with the number decreasing as the percentage of consumers declines.
The model was tested on a dataset composed by 5 different types of Users:
\begin{itemize}
    \item \texttt{Type 0}: Prosumer with a photovoltaic (PV) system.
    \item \texttt{Type 1}: Prosumer with a PV system and an energy storage system (ESS).
    \item \texttt{Type 2}: Prosumer with a PV system integrated with an electric vehicle (EV).
    \item \texttt{Type 3}: Consumer with an electric vehicle (EV) without PV integration.
    \item \texttt{Type 4}: Consumer, labeled as "Vanilla," represent a standard or baseline consumer type with typical energy consumption, not associated with specialized systems like EVs or PV systems.
\end{itemize}

It is important to consider that, in the small community where the dataset was trained and tested, only these 4 types of energy produced was considered, which could introduce a bias during testing with larger communities that also have wind energy sources (which represents the second most adopted energy source in Denmark after photovoltaic energy). This bias especially impacts the winter and spring months, during which there are strong wind gusts but a general lack of photovoltaic energy.
It is also important to note another strong bias due to the location from which the dataset was obtained: being a relatively cold country like Denmark, the winter months have the coldest temperatures, while the spring and summer months, unlike other countries in, for example, Southern Europe, do not experience particularly high heat peaks, making the model not generalizable to all countries, as the temperature variability is too high.

Therefore, the final model forecasts the energy surplus accumulated by the community for each hour of the year (for visualization purposes, a value every 404 hours is shown in the line chart). The ideal value is obtained by subtracting the imported energy from the produced energy after performing an exponentially weighted mean of every imported/exported form of energy. The simulation, therefore, provides a value for each hour of the year based on the composition of the LEC, indicating the current energy availability. This value can be positive in the case of a surplus of produced energy, which is a trend observed during the warm summer and spring months, and negative in the case of a surplus of consumed energy, despite the produced energy, which is a trend observed during the cold autumn and winter months.

\section{Conclusion}
\label{conclusions}

In conclusion, the application of FL to energy communities represents a significant advancement in the field of energy management and forecasting. By enabling decentralized, privacy-preserving model training, FL addresses key challenges associated with the sensitive and distributed nature of energy data within these communities. This approach not only enhances the accuracy of energy consumption and production forecasts by leveraging diverse local data but also ensures compliance with data privacy regulations, a critical concern in the modern energy landscape.

The potential of FL extends beyond mere forecasting; it offers a pathway towards more resilient and adaptive energy systems. As energy communities continue to evolve, becoming increasingly decentralized and interconnected, FL could play a pivotal role in optimizing energy distribution, enhancing self-sufficiency, and supporting the broader goals of sustainability and carbon neutrality. Future research should focus on refining FL algorithms to improve their efficiency and scalability in large, heterogeneous energy communities, as well as exploring the integration of FL with other emerging technologies, such as blockchain and edge computing, to further enhance its capabilities.

In addition we plan to develop applications and conduct more experiments  applying the above forecasting mechanism to LEC optimization and management strategies.\\

\noindent {\bf Acknowledgement.} Work supported by the NextGenerationEU project: Ecosystem for Sustainable Transition in Emilia-Romagna (Ecosister) CUP: B33D21019790006 - PNRR - Missione 4 Componente 2 Investimento 1.5 - Spoke 4.

%
%
%

\end{document}